\documentclass[conference]{IEEEtran}

\ifCLASSINFOpdf

\else

\fi
\usepackage{hyperref}
\usepackage{booktabs}
\usepackage{multirow}
\usepackage[table]{xcolor}
\usepackage{enumitem}
\usepackage{graphicx}
\usepackage{amsmath}
\usepackage{amssymb}
\usepackage{fontawesome}
\begin{document}
%

\title{TD-RD: A Top-Down Benchmark with Real-Time Framework for Road Damage Detection}


\author{
    Xi Xiao\textsuperscript{1},\qquad Zhengji Li\textsuperscript{1}, \qquad Wentao Wang\textsuperscript{1}, \qquad Jiacheng Xie\textsuperscript{1}, \qquad \\
    Houjie Lin\textsuperscript{1}, \qquad Swalpa Kumar Roy\textsuperscript{2}, \qquad Tianyang Wang\textsuperscript{1}\textsuperscript{\faEnvelope}, \qquad Min Xu\textsuperscript{3,4}\textsuperscript{\faEnvelope} \\
    \textsuperscript{1}University of Alabama at Birmingham, \qquad
    \textsuperscript{2}Alipurduar Government Engineering and Management College, \qquad \\
    \textsuperscript{3}MBZUAI, \qquad \textsuperscript{4}Carnegie Mellon University \\
    \vspace{0.5em} 
    \textsuperscript{\faEnvelope} Corresponding authors:(\href{mailto:mxu1@cs.cmu.edu}{mailto:mxu1@cs.cmu.edu}), (\href{mailto:tw2@uab.edu}{tw2@uab.edu})
}


%


\maketitle

\begin{abstract}
Object detection has witnessed remarkable advancements over the past decade, largely driven by breakthroughs in deep learning and the proliferation of large-scale datasets. However, the domain of road damage detection remains relatively underexplored, despite its critical significance for applications such as infrastructure maintenance and road safety. This paper addresses this gap by introducing a novel top-down benchmark that offers a complementary perspective to existing datasets, specifically tailored for road damage detection. Our proposed Top-Down Road Damage Detection Dataset (TD-RD) includes three primary categories of road damage—cracks, potholes, and patches—captured from an top-down viewpoint. The dataset consists of 7,088 high-resolution images, encompassing 12,882 annotated instances of road damage. Additionally, we present a novel real-time object detection framework, TD-YOLOV10, designed to handle the unique challenges posed by the TD-RD dataset. Comparative studies with state-of-the-art models demonstrate competitive baseline results. By releasing TD-RD, we aim to accelerate research in this crucial area. A sample of the dataset will be made publicly available upon the paper's acceptance.
\end{abstract}


%
\IEEEpeerreviewmaketitle

\section{Introduction}

Timely maintenance of road damage can greatly enhance transportation safety, making the development of advanced technologies for efficient detection crucial. Thanks to advancements in deep learning and the availability of large-scale datasets over the past decade, target detection has made remarkable progress. Several benchmarks have been introduced to support road damage detection~\cite{arya2022crowdsensingbasedroaddamagedetection,li2024cycleyoloefficientrobustframework,2020rebalance,9904187,ji2024advancingoutofdistributiondetectiondata}. However, existing benchmarks often lack multi-view information. The challenge of detecting and localizing road damage requires robust approaches that incorporate multi-view data. To address this gap, we propose the Top-Down Road Damage Dataset (TD-RD), which offers a supplemental camera angle to existing benchmarks. TD-RD consists of 7,088 high-resolution pavement images with clearly identifiable damage instances, selected from 214,872 self-acquired images through rigorous filtering and cleaning processes. The damage instances are classified into three categories: cracks, potholes, and repairs, ensuring balanced representation across classifications. In addition to the TD-RD dataset, we introduce a new real-time detection model, TD-YOLOV10, designed to leverage top-down data for enhanced robustness and accuracy in detection. TD-YOLOV10 incorporates a self-attention\cite{han2022survey} framework and a Multi-Scale Attention with Positional Squeeze-and-Excitation (MAPSE) mechanism to enhance the detection and recognition of road pavement images. MAPSE combines multi-scale and dual attention mechanisms to effectively fuse features at different scales\cite{chen2024features,xiao2024hgtdpdtahybridgraphtransformerdynamic,10706324}, addressing the limitations of traditional convolutional networks by improving complex feature representation. This architecture captures and fuses feature representations across various scales, striking a balance between accuracy and efficiency. We compare TD-YOLOV10 with mainstream detectors using TD-RD and two other public road pavement datasets: CNRDD and CRDDC’22. Our framework demonstrates significant competitiveness in both detection accuracy and efficiency.

In summary, our contributions are as follows: 1) We present the first attempt to explore road damage detection from a top-down perspective, offering a complementary view to existing datasets. 2) We introduce TD-RD, the first benchmark specifically designed for detecting road damage from a top-down perspective. It includes three primary categories of road damage: cracks, potholes, and patches. 3) We developed TD-YOLOV10, a real-time framework for detecting road damage from a top-down perspective. Extensive experiments demonstrated the competitive performance of TD-YOLOV10 on TD-RD, CNRDD, and CRDDC’22.

\begin{figure}[htbp]
\centering
\includegraphics[width=\linewidth]{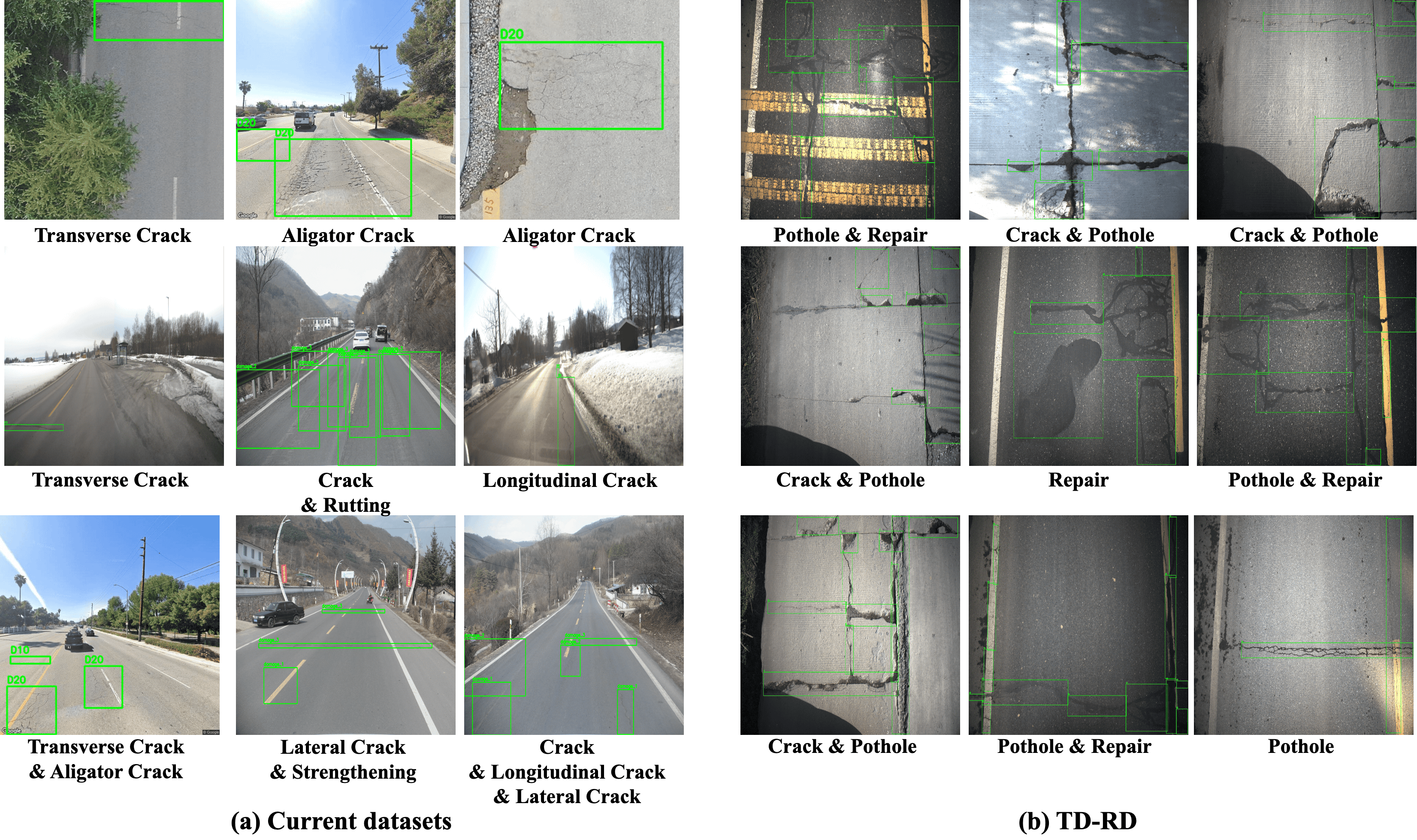}
\caption{Annotated Samples from current datasets and TD-RD, which contains annotation distribution and categories.}
\label{Dataset}
\end{figure}

\begin{figure*}[!htbp]
\centering

\includegraphics[width=15cm]{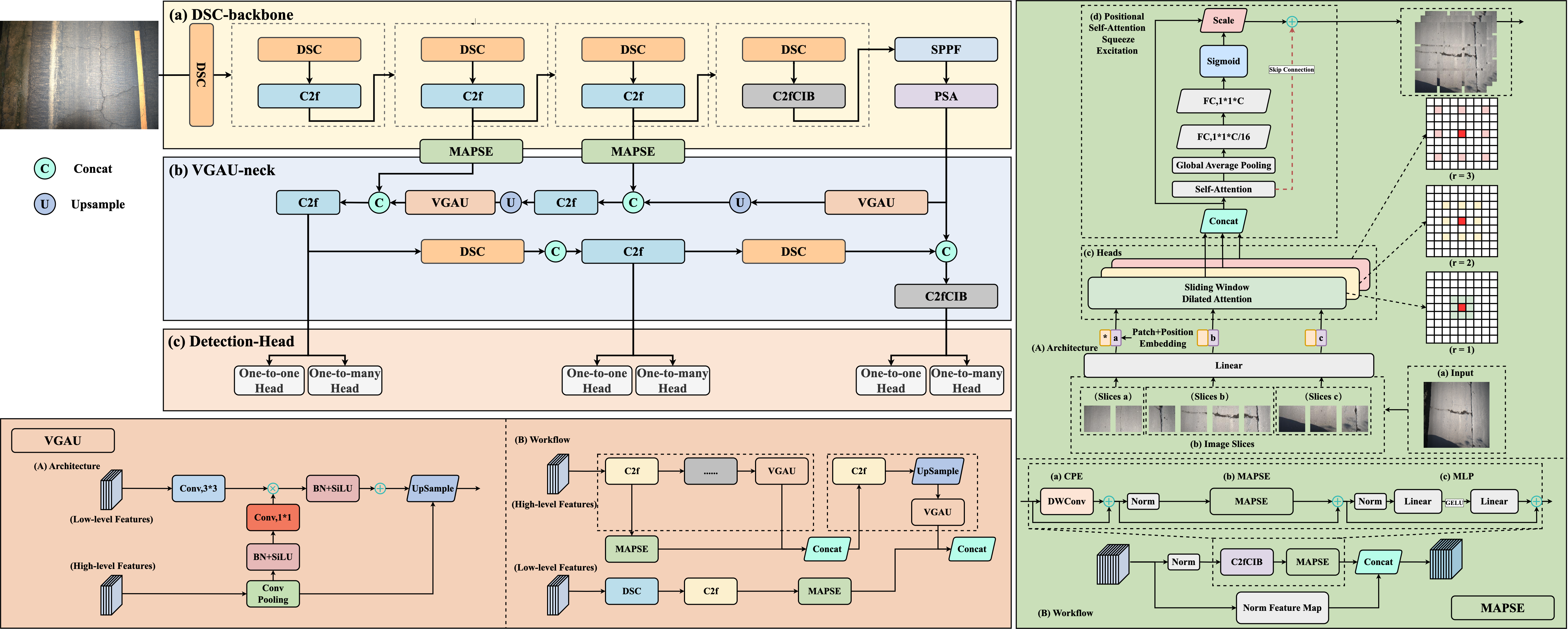}
\caption{Visualization of model architecuture, which contains the architecture\textbf{(A)} and workflow\textbf{(B)} of MAPSE module and VGAU module.}
\label{Dataset1}
\end{figure*}

\section{Benchmark}

A top-down camera viewpoint is crucial for accurate and comprehensive modeling of road surface conditions. This perspective, when combined with information from additional sources, offers a holistic approach to capturing the intricacies of road damage. To this end, we introduce TD-RD, the first fully top-down road damage dataset designed for detection tasks. The dataset was curated using high-speed, high-definition vehicle-mounted cameras operated by professionals. From an initial pool of 214,872 images, a rigorous selection process yielded 7,088 high-quality images, each with a resolution of 3840 × 2160 pixels, ensuring exceptional clarity and detail. All images were meticulously labeled according to a framework co-developed by experts from the geology and highway engineering sectors. This labeling process identified critical road damage types, including cracks, potholes, and repairs. TD-RD is balanced across these three major categories, featuring 10,342 instances of cracks, 8,763 potholes, and 10,457 repairs. This balanced distribution ensures that models trained on TD-RD achieve robust generalization across diverse road damage types, enhancing their accuracy and reliability in real-world detection tasks.

\section{Method}

The proposed TD-YOLOv10 has three core components: Backbone, Neck, and Detection Head. In the \textbf{Backbone}, we integrate the Multi-Scale Attention with Positional Squeeze-and-Excitation (MAPSE) module to capture long-range contextual dependencies across multiple scales while preserving fine-grained feature details. Depthwise Separable Convolution (DSC) layers, as described by \cite{li2022efficient}, reduce computational complexity, accelerating inference and enhancing the model's generalization and deployment efficiency. The \textbf{Neck} incorporates a Visual Global Attention Upsampling (VGAU) module, which facilitates the extraction of global contextual information, aiding in the refinement of crucial features for accurate category classification. The synergy between VGAU and MAPSE enables the model to capture and retain both global and local information, significantly improving detection performance. Finally, the \textbf{Detection Head} adopts a multi-scale prediction strategy, predicting bounding boxes at three distinct scales, thereby enabling the accurate detection of objects of varying sizes within pavement damage images.

\subsection{MAPSE-based Feature Extraction}

Efficiently capturing global information and long-distance relationships in road pavement images with complex backgrounds is crucial for enhancing semantic discriminability and reducing category confusion. To address this, we introduce the Multi-Scale Attention with Positional Squeeze-and-Excitation (MAPSE) module, inspired by Visual Transformers\cite{kai2023asurvey}. The MAPSE module integrates multi-scale feature fusion within the YOLOv10 architecture. The MAPSE module starts by dividing the input image into slices, with each slice undergoing patch and position embedding to retain spatial context. Feature maps are then processed through a multi-head attention mechanism with varying dilation rates to capture multi-scale features. This is represented as:

\begin{equation}
\text{Dilation}(x, r) = \sum_{i=1}^{k} w_i x_{i \cdot r}
\end{equation}

where $\text{Dilation}(x, r)$ represents the dilation operation on input $x$ with dilation rate $r$, $w_i$ are the weights, $x_{i \cdot r}$ is the input feature map at position $i$ dilated by rate $r$. The dilation rates are set as $r_1 = 1$, $r_2 = 2$, and $r_3 = 4$ to capture different scales. After self-attention, global average pooling captures channel-wise global information, and fully connected layers learn dependencies:

\begin{equation}
F_{\text{recalibrated}} = \sigma(W_2 \cdot \text{ReLU}(W_1 \cdot F_{\text{global}}))
\end{equation}

where $F_{\text{recalibrated}}$ is the recalibrated feature map, $F_{\text{global}}$ is the global average pooled feature, $W_1$ and $W_2$ are learnable weight matrices, $\text{ReLU}$ is the Rectified Linear Unit activation function, and $\sigma$ is the sigmoid activation function. A skip connection bypasses the MAPSE module, adding input features directly to the recalibrated features to preserve original information and improve gradient flow during training:

\begin{equation}
F_{\text{final}} = F_{\text{recalibrated}} + F_{\text{input}}
\end{equation}

where $F_{\text{final}}$ is the final output feature map, $F_{\text{input}}$ is the original input feature map, and the addition operation represents the skip connection.

The structure and flow of the MAPSE module are illustrated in Fig.~\ref{Dataset1}. The MAPSE module combines Depthwise Convolution (DWConv), Multi-scale Sliding Window Expansion Attention (SWDA), and a Multi-Layer Perceptron (MLP), enabling the detector to retain underlying semantic information and enrich global feature extraction at the neck layer. By integrating the MAPSE module, our architecture effectively captures long-range dependencies and contextual information, which is crucial for distinguishing complex damage patterns in road images. The innovative use of multi-scale attention and positional squeeze-and-excitation mechanisms directly enhances the model’s performance in road pavement damage detection, as demonstrated by our experimental results.

\subsection{Visual Global Attentional Upsampling}

Efficient multi-scale feature representation is crucial for detecting road damage of varying sizes. Traditional convolutional networks often fail to represent these features effectively. To overcome this limitation, we developed the Visual Global Attentional Upsampling (VGAU) module, which enhances multi-scale feature representation without significantly increasing computational costs. As shown in Fig.~\ref{Dataset1}, the VGAU module uses global average pooling to incorporate global context information for semantic discrimination. A 3×3 convolution compresses the number of channels in low-dimensional features, refining feature representation by integrating high-dimensional features. The high-dimensional features are fused with weighted low-dimensional features, followed by an upsampling operation to produce the final feature representation. Unlike GAU\cite{li2018pyramidattentionnetworksemantic}, VGAU employs the SiLU\cite{stefan2018sigmoid} activation function and sums high-dimensional and low-dimensional features, ensuring efficient and accurate multi-scale feature representation.

\section{Experiments}

\subsection{Datasets}

TD-RD comprises 7,088 images with 3840×2160 pixels containing 12,882 annotated damage instances. The dataset offers balanced representation across three key damage categories: cracks, potholes, and repairs, each accompanied by meticulously labeled bounding boxes for precise detection tasks. CRDDC’22 consists of 38,387 images with resolutions varying from 512×512 to 4040×2035 pixels, capturing a diverse range of road damage types, including longitudinal, transverse, and alligator cracks and potholes. CNRDD comprises 4,318 images with resolutions up to 1600×1200 pixels, encompassing a wide variety of road damage types, including cracks, longitudinal cracks, lateral cracks, subsidence, rutting, potholes, surface looseness, surface strengthening, and uncertain damage categories.

\subsection{Evaluation Metrics}
\label{evaluation_metrics}
Following established evaluation protocols~\cite{Abdelraouf2022Vision, Gochoo_2023_CVPR}, we utilize multiple metrics to assess the performance of our proposed detector. Key metrics include Mean Average Precision (mAP\%) and Precision (Pre\%), which evaluate detection accuracy. Specifically, mAP is calculated based on Average Precision (AP) at varying Intersections over Union (IoU) thresholds, ranging from 0.50 to 0.95. Precision (Pre\%) is measured at a confidence threshold 0.5 and serves as the primary ranking metric. Additionally, we incorporate Floating Point Operations (FLOPs) and Frames Per Second (FPS) to quantify the computational efficiency of the models.

\subsection{Implementation details}

All experiments were conducted using the PyTorch framework on an Nvidia A100 GPU. The input images were resized to 512×512 pixels. The model was trained using the SGD optimizer with an initial learning rate of 0.01, which was reduced to 0.002. We used a weight decay factor of 0.0005 and a momentum parameter of 0.8. The training was performed over 300 epochs with a batch size of 32.

\section{Results}

\begin{table*}[!htbp]
\caption{Comparisons with state-of-the-arts on Effectiveness. The best results are highlighted in \textbf{bold}. The second best results are highlighted in {\color{red}{red}}, while the third best results are highlighted in {\color{blue}{blue}}}
\centering
\scalebox{0.7}{
\setlength{\tabcolsep}{0.15pt} 
\small 
\begin{tabular*}{\textwidth}{@{\extracolsep{\fill}}lcccccccccccc@{}}
\toprule
\multirow{2}{*}{Model} & \multicolumn{4}{c}{TD-RD} & \multicolumn{4}{c}{CNRDD} & \multicolumn{4}{c}{CRDDC'22}\\ 
\cmidrule(ll){2-5} \cmidrule(ll){6-9} \cmidrule(ll){10-13}
& mAP($\%$) $\uparrow$ & Pre($\%$) $\uparrow$ & FLOPs(G) & FPS & mAP($\%$) $\uparrow$ & Pre($\%$) $\uparrow$ & FLOPs(G) & FPS & mAP($\%$) $\uparrow$ & Pre($\%$) $\uparrow$ & FLOPs(G) & FPS \\
\midrule
YOLOv5-n & 81.4 & 79.8 & \textbf{4.10} & 139 & 21.4 & 33.7 & \textbf{4.10} & 139 & 41.4 & 44.7 & \textbf{4.10} & 139  \\
YOLOv5-s & 85.6 & 84.6 & 15.8 & 111 & 22.5 & 30.7 & 15.8 & 111 & 42.1 & 46.4 & 15.8 & 111  \\
YOLOv6-n~\cite{li2022yolov6singlestageobjectdetection} \raisebox{0.5pt}{\color{gray}\scriptsize [arXiv'22]} & 78.3 & 76.9 & 11.4 & 123 & 21.4 & 31.8 & 11.4 & 123 & 42.1 & 46.4 & 11.4 & 123  \\
YOLOv6-s~\cite{li2022yolov6singlestageobjectdetection} \raisebox{0.5pt}{\color{gray}\scriptsize [arXiv'22]} & 83.0 & 82.5 & 45.3 & 81 & 24.6 & 33.8 & 45.3 & 81 & 42.4 & 46.0 & 45.3 & 81 \\
YOLOv7-ti~\cite{Wang_2023_CVPR} \raisebox{0.5pt}{\color{gray}\scriptsize [CVPR'23]} & 84.5 & 85.7 & 13.2 & {\color{blue}{294}} & 25.3 & 33.5 & 13.2 & {\color{blue}{294}} & 46.2 & 49.8 & 13.2 & {\color{blue}{294}}  \\
YOLOv8-n~\cite{reis2024realtimeflyingobjectdetection} \raisebox{0.5pt}{\color{gray}\scriptsize [arXiv'24]} & 82.2 & 81.9 & {\color{red}{8.2}} & \textbf{385} & 27.6 & 38.4 & {\color{red}{8.2}} & \textbf{385} & 46.0 & 48.5 & {\color{red}{8.2}} & \textbf{385}  \\
YOLOv8-s~\cite{reis2024realtimeflyingobjectdetection} \raisebox{0.5pt}{\color{gray}\scriptsize [arXiv'24]} & 85.1 & 86.0 & 28.4 & 333 & 27.6 & 38.4 & 28.4 & 333 & 46.0 & 48.5 & 28.4 & 333  \\
YOLOv9-s~\cite{wang2024yolov9learningwantlearn} \raisebox{0.5pt}{\color{gray}\scriptsize [arXiv'24]} & 85.2 & \textbf{88.6} & 30.3 & 172 & {\color{red}{29.5}} & 37.4 & 30.3 & 172 & {\color{blue}{47.4}} & 49.7 & 30.3 & 172  \\
YOLOv10-n~\cite{wang2024yolov10realtimeendtoendobject} \raisebox{0.5pt}{\color{gray}\scriptsize [arXiv'24]} & 82.3 & 81.4 & {\color{blue}{8.22}} & {\color{red}{357}} & 28.1 & 35.9 & {\color{blue}{8.22}} & {\color{red}{357}} & 46.5 & 48.3 & {\color{blue}{8.22}} & {\color{red}{357}}  \\
YOLOv10-s~\cite{wang2024yolov10realtimeendtoendobject} \raisebox{0.5pt}{\color{gray}\scriptsize [arXiv'24]} & 85.0 & 82.2 & 24.5 & 286 & 28.8 & 39.4 & 24.5 & 286 & 47.3 & \textbf{57.3} & 24.5 & 286  \\
YOLOS-ti~\cite{fang2021looksequencerethinkingtransformer} \raisebox{0.5pt}{\color{gray}\scriptsize [arXiv'21]} & 80.8 & 80.4 & 21 & 116 & 21.3 & 30.1 & 21 & 116 & 45.3 & {\color{blue}{52.0}} & 24.5 & 286  \\
YOLOS-s~\cite{fang2021looksequencerethinkingtransformer} \raisebox{0.5pt}{\color{gray}\scriptsize [arXiv'21]}  & 84.7 & 83.2 & 179 & 54 & 23.6 & 36.4 & 179 & 54 & 46.8 & 49.4 & 179 & 54  \\
\midrule
PP-PicoDet~\cite{yu2021pppicodetbetterrealtimeobject} \raisebox{0.5pt}{\color{gray}\scriptsize [arXiv'21]} & 85.6 & 83.4 & 8.9 & 196 & 22.4 & 31.7 & 8.9 & 196 & 47.0 & 48.0 & 8.9 & 196  \\
RT-DERT~\cite{zhao2024detrsbeatyolosrealtime} \raisebox{0.5pt}{\color{gray}\scriptsize [CVPR'23]} & {\color{red}{87.7}} & {\color{blue}{87.7}} & 60 & 159 & {\color{blue}{29.4}} & {\color{blue}{39.5}} & 60 & 159 & \textbf{48.6} & 51.7 & 60 & 159  \\
Lite-DERT~\cite{Yu_2021_CVPR} \raisebox{0.5pt}{\color{gray}\scriptsize [CVPR'21]} & {\color{blue}{86.1}} & 85.2 & 151 & 75 & 26.3 & 33.0 & 151 & 75 & 45.3 & 48.9 & 151 & 75  \\
\midrule
FR-CNN~\cite{NIPS2015_14bfa6bb} \raisebox{0.5pt}{\color{gray}\scriptsize [NIPS'15]}  & 74.6 & 76.6 & 94.3 & 10 & 20.3 & 36.3 & 94.3 & 10 & 39.9 & 46.1 & 94.3 & 10  \\
SSD-VGG16~\cite{Liu_2016} \raisebox{0.5pt}{\color{gray}\scriptsize [ECCV'16]} & 66.5 & 71.1 & 60.9 & 14 & 18.5 & \textbf{49.9} & 60.9 & 14 & 38.7 & 46.2 & 60.9 & 14  \\
\midrule
TD-YOLOV10 \raisebox{0.5pt}{\color{gray}\scriptsize [Ours]} & \textbf{88.1} & {\color{red}{88.5}} & 31.8 & 200 & \textbf{37.0} & {\color{red}{46.0}} & 31.8 & 200 &  {\color{red}{48.0}} & {\color{red}{54.1}} & 31.8 & 200  \\
\bottomrule
\end{tabular*}}
\label{Experimental_Results}
\end{table*}

\setlength{\tabcolsep}{0.15cm} 
\begin{table*}[!htbp]
\caption{Ablation studies conducted on each component. The best results are highlighted in \textbf{bold}}
\label{tab:ablation}
\centering
\scalebox{0.7}{
\setlength{\tabcolsep}{2.5pt} 
\small 
\begin{tabular*}{\textwidth}{@{\extracolsep{\fill}}ccccccccccccccc@{}}
\toprule
\multicolumn{3}{c}{Module Settings} & \multicolumn{4}{c}{TD-RD} & \multicolumn{4}{c}{CNRDD} & \multicolumn{4}{c}{CRDDC'22} \\ 
\cmidrule{1-3} \cmidrule{4-7} \cmidrule{8-11} \cmidrule{12-15}
DSC & MAPSE & VGAU & mAP($\%$) & Pre($\%$) & FLOPs(G) & FPS & mAP($\%$) & Pre($\%$) & FLOPs(G) & FPS & mAP($\%$) & Pre($\%$) & FLOPs(G) & FPS \\ 
\midrule
 & $\checkmark$ &  & 86.2 & 85.2 & 27.3 & 260 & 32.0 & 41.5 & 27.3 & 260 & 43.0 & 49.5 & 27.3 & 260 \\ 
$\checkmark$ & $\checkmark$ &  & 86.8 & 86.8 & 28.5 & 246 & 33.8 & 43.0 & 28.5 & 246 & 44.8 & 51.0 & 28.5 & 246 \\ 
 & $\checkmark$ & $\checkmark$ & 87.5 & 87.9 & 30.0 & 218 & 35.5 & 44.5 & 30.0 & 218 & 46.5 & 52.5 & 30.0 & 218 \\ 
$\checkmark$ & $\checkmark$ & $\checkmark$ & \textbf{88.1} & \textbf{88.5} & \textbf{31.8} & \textbf{200} & \textbf{37.0} & \textbf{46.0} & \textbf{31.8} & \textbf{200} & \textbf{48.0} & \textbf{54.1} & \textbf{31.8} & \textbf{200} \\ 
\bottomrule
\end{tabular*}}
\end{table*}

\subsection{Comparison with existing works}

\subsubsection{Results on TD-RD Dataset}

Table~\ref{Experimental_Results} presents a comparative analysis of TD-YOLOV10 against state-of-the-art methods on the TD-RD dataset. Our proposed TD-YOLOV10 achieves an impressive mAP of 88.1\% and a Precision of 88.5\%, significantly surpassing multiple YOLO variants and other SOTA models. In particular, TD-YOLOV10 outperforms YOLOv8-s with a notable gain of +3.0\% mAP and +2.5\% Precision, and RT-DERT with improvements of +0.4\% mAP and +0.8\% Precision. Furthermore, the model sustains a high frame rate of 200 FPS, ensuring its suitability for real-time applications.

\subsubsection{Results on CNRDD Dataset}

Evaluating TD-YOLOV10 on the CNRDD dataset further highlights its robustness. As detailed in Table~\ref{Experimental_Results}, TD-YOLOV10 achieves an mAP of 37.0\% and a Precision of 46.0\%, outperforming competing models such as YOLOS-s (23.6\% mAP) and Lite-DERT (26.3\% mAP). The model's high processing speed, sustaining 200 FPS, underscores its efficiency, making it well-suited for real-time road damage detection systems.

\subsubsection{Results on CRDDC’22 Dataset:}

On the CRDDC’22 dataset, TD-YOLOV10 achieves an mAP of 48.0\% and a Precision of 54.1\%, surpassing state-of-the-art models such as YOLOv10-s (47.3\% mAP) and PP-PicoDet (47.0\% mAP). With a notably lower computational cost of 31.8 GFLOPs and an impressive inference speed of 200 FPS, TD-YOLOV10 showcases its strong balance between accuracy and efficiency, making it well-suited for real-time road damage detection. The results on the CRDDC’22 dataset, which features a broader range of crack types and varying image resolutions, underscore the model’s robustness and adaptability in addressing diverse and complex road damage scenarios.

\subsection{Ablations}

\subsubsection{Effectiveness of MAPSE}

Integrating MAPSE led to substantial improvements in both mean Average Precision (mAP) and Precision across all evaluated datasets. The enhanced performance, attributed to MAPSE, stems from its ability to effectively capture long-range dependencies and contextual information. On the TD-RD dataset, the model achieved an impressive mAP of 86.2\% and Precision of 85.2\%, with a computational cost of 27.3 GFLOPs. For the CNRDD dataset, mAP reached 32.0\% with a Precision of 41.5\%, while on the CRDDC'22 dataset, mAP and Precision were 43.0\% and 49.5\%, respectively.

\subsubsection{Impact of Combining DSC and MAPSE}

The integration of DSC with MAPSE not only enhances computational efficiency but also sustains high detection accuracy across various benchmarks. On the TD-RD dataset, this combination resulted in an improved mAP of 86.8\% and Precision of 86.8\%, with a computational overhead of 28.5 GFLOPs. For the CNRDD dataset, the model achieved a mAP of 33.8\% and a Precision of 43.0\%. Meanwhile, on the CRDDC’22 dataset, which presents a variety of complex scenarios, mAP and Precision reached 44.8\% and 51.0\%, respectively.

\subsubsection{Contribution of VGAU}

The incorporation of VGAU significantly bolstered the model's capacity to capture fine-grained and contextual information, leading to notable performance gains. On the TD-RD dataset, mAP increased to 87.5\% and Precision to 87.9\%, with a computational complexity of 30.0 GFLOPs. For the CNRDD dataset, mAP improved to 35.5\%, with Precision rising to 44.5\%. On the challenging CRDDC'22 dataset, mAP and Precision reached 46.5\% and 52.5\%, respectively.

\section{Conclusion}

In this paper, we introduce TD-RD, the first benchmark designed to advance road damage detection by incorporating a top-down perspective information. TD-RD spans a diverse array of road types and damage scenarios, consisting of 7,088 high-resolution images and 12,882 annotated damage instances, categorized with standard bounding boxes and damage types. Our experimental results highlight the limitations of existing methods when confronted with the road damage detection from a top-down view. To this end, we propose TD-YOLOv10, an efficient detector based on the YOLOv10 framework, enhanced with the Multi-Scale Attention with Positional Squeeze-and-Excitation (MAPSE) module and the Vision-based Global Attention Upsampling (VGAU) module. These improvements substantially bolster the model's capacity to capture global context and discern fine-grained damage features, resulting in superior performance on road pavement damage detection tasks. By releasing TD-RD, we aim to promote further research in the field of road disease detection, emphasizing the key role of top-down view information in advancing the field of road disease detection.



\clearpage

\bibliographystyle{ieee_fullname} 
\bibliography{egbib} 

\end{document}